\documentclass[10pt,twocolumn,letterpaper]{article}

\usepackage{cvpr}              
\definecolor{cvprblue}{rgb}{0.21,0.49,0.74}
\usepackage[pagebackref,breaklinks,colorlinks,allcolors=cvprblue]{hyperref}
\usepackage{multirow}

\definecolor{Red}{RGB}{192, 0, 0}
\definecolor{Blue}{RGB}{12, 114, 186}

\def\paperID{*****} 
\def\confName{CVPR}
\def\confYear{2026}
\newcommand\blfootnote[1]{%
  \begingroup
  \renewcommand\thefootnote{}\footnote{#1}%
  \addtocounter{footnote}{-1}%
  \endgroup
}
\title{EasyVFX: Frequency-Driven Decoupling for Resource-Efficient VFX Generation}

\author{
Yue Ma\textsuperscript{1} \quad
Xu Ye\textsuperscript{1} \quad
Qinghe Wang\textsuperscript{2\dag} \quad
Yucheng Wang\textsuperscript{1} \quad
Hongyu Liu\textsuperscript{1} \quad
Yinhan Zhang\textsuperscript{1} \\
Xinyu Wang\textsuperscript{3} \quad
Yuanpeng Che \quad
Shanhui Mo \quad
Paul Liang\textsuperscript{5} \quad
Fangneng Zhan\textsuperscript{1\dag} \quad
Qifeng Chen\textsuperscript{1}
\\
\\
\textsuperscript{1}HKUST,  \quad
\textsuperscript{2}DUT, \quad
\textsuperscript{3}THU, \quad
\textsuperscript{5}MIT
\\
\\
Project Page: \url{https://easy-vfx.github.io/}
}
\begin{document}


\twocolumn[{
\renewcommand\twocolumn[1][]{#1}%
\maketitle

\begin{center}
  \centering
  \includegraphics[width=\linewidth]{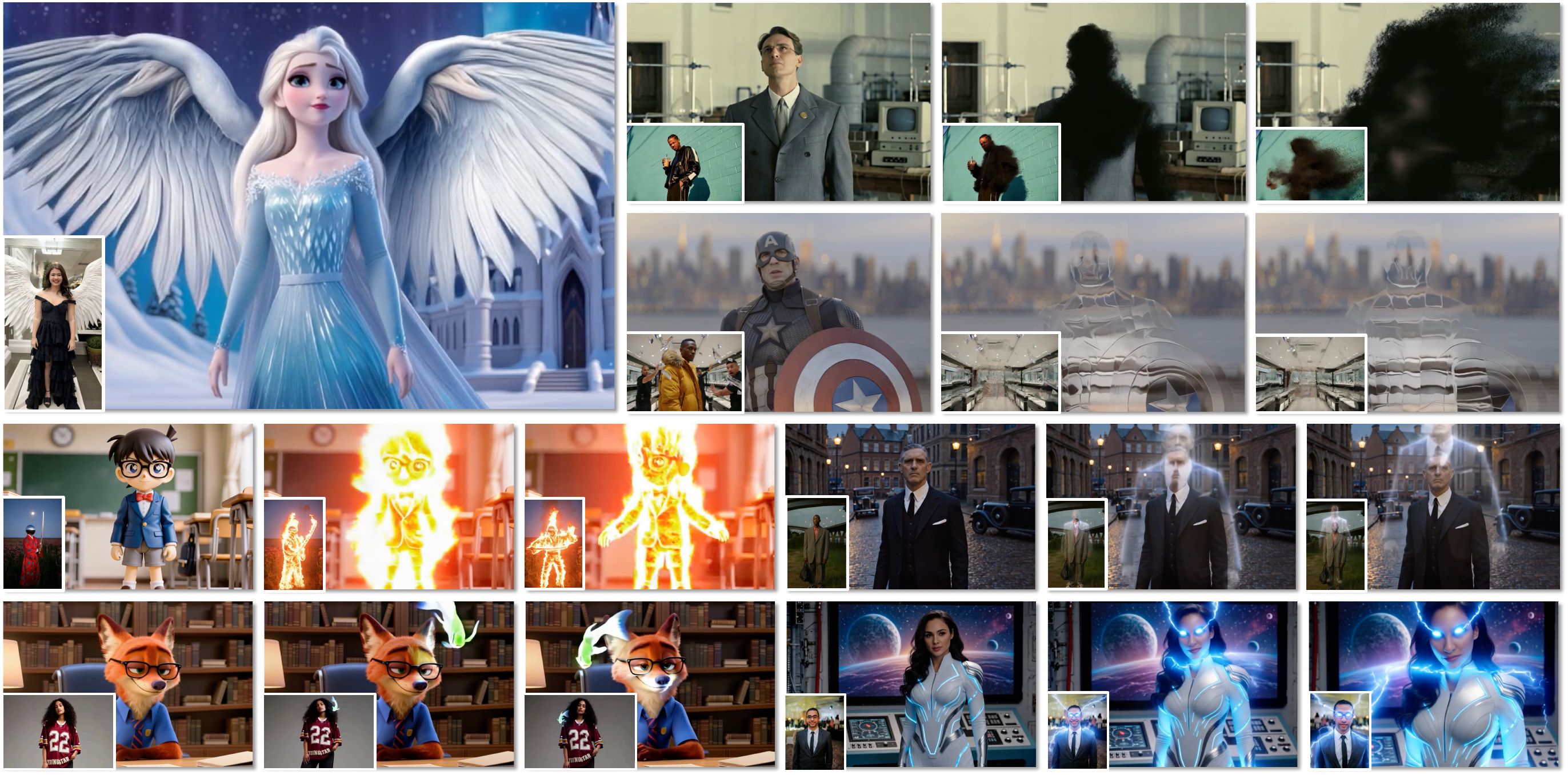}
  \captionof{figure}{   \textbf{Showcase of visual effect generation} produced by our proposed \textbf{EasyVFX}, which allows users to perform high-fidelity visual effect generation while maintaining coherent structures and temporal consistency following the reference video.
    The reference videos are shown in the lower left corner of the generated results, respectively.
    \textcolor{black}{The input images are synthetic images produced by publicly available LoRA models~\cite{civitai}.}
  }
\end{center}
}]
\blfootnote{* Equal contribution.}
\blfootnote{\textsuperscript{\dag} Corresponding author.}

\begin{abstract}
Generating high-fidelity visual effects (VFX) typically demands massive datasets and prohibitive computational power due to the intricate coupling of spatial textures and temporal dynamics. In this paper, we introduce EasyVFX, a resource-efficient framework that achieves realistic VFX synthesis under stringent constraints. Our core philosophy lies in frequency-domain decomposition: we observe that the complexity of VFX can be significantly mitigated by decoupling high-frequency components, which represent intricate spatial appearances, from low-frequency components that encapsulate global motion dynamics. This spectral disentanglement transforms a high-dimensional learning problem into manageable sub-tasks, thereby lowering the optimization barrier and reducing data dependency. Building upon this insight, we propose a two-stage training paradigm. First, we design a Frequency-aware Mixture-of-Experts (Freq-MoE) architecture. By utilizing a soft routing mechanism, our model assigns specialized experts to distinct spectral bands, enabling them to cultivate robust priors for appearance and motion dynamics. This specialization allows the model to acquire foundational VFX knowledge with fewer GPU resources. Second, we introduce a Test-Time Training strategy powered by a novel Frequency-constraint Loss. This allows the pre-trained model to swiftly adapt to specific, unseen effects through localized optimizations, requiring only about 100 steps on a single GPU. Experimental results demonstrate that EasyVFX produces structurally consistent and visually stunning effects, proving that frequency-aware learning is a key catalyst for democratizing professional-grade VFX.
\end{abstract}

\section{Introduction}

\begin{figure*}[t]
    \centering
    \includegraphics[width=\textwidth]{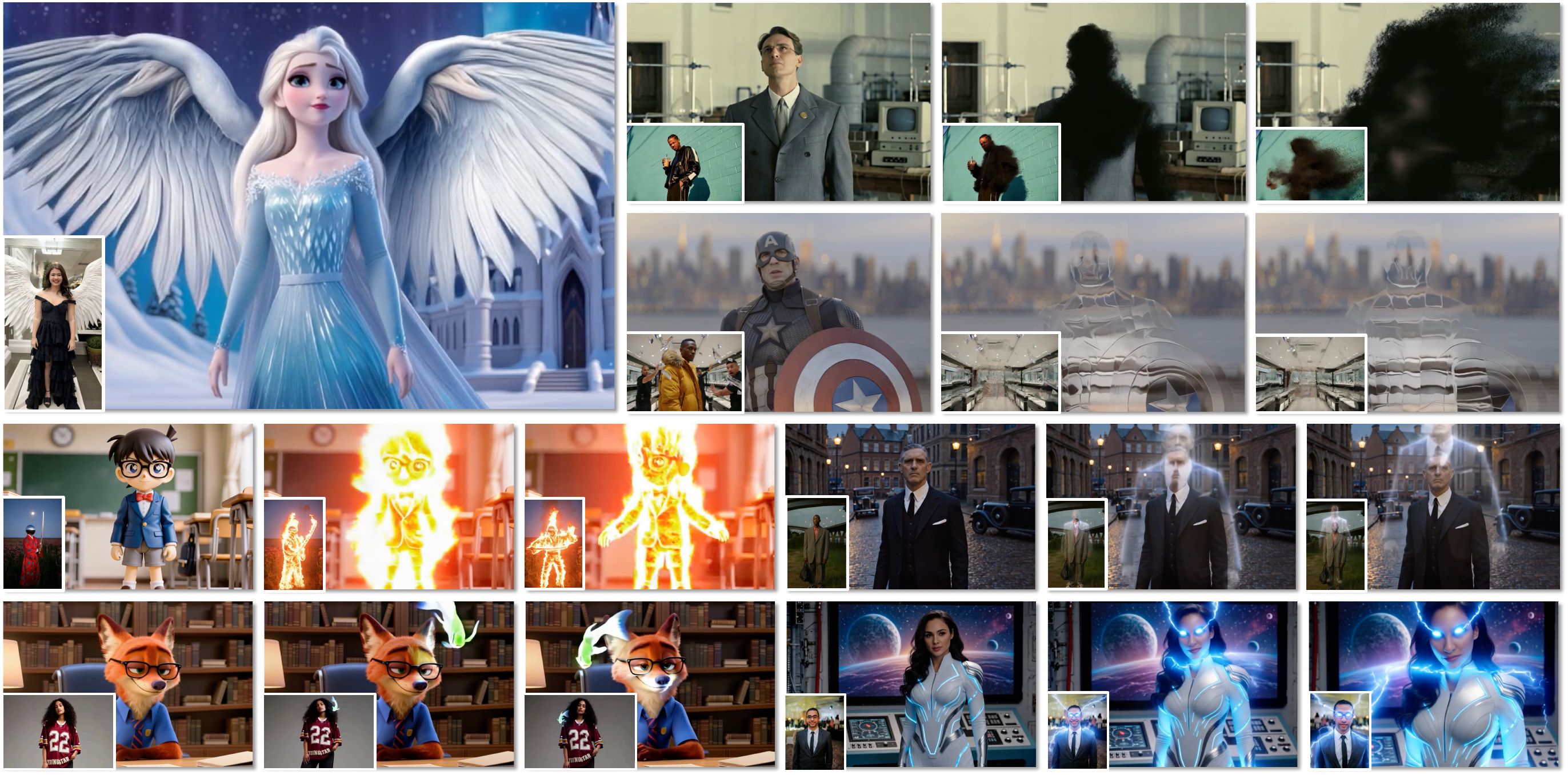}
    \caption{
    \textbf{Showcase of visual effect generation} produced by our proposed \textbf{EasyVFX}, which allows users to perform high-fidelity visual effect generation while maintaining coherent structures and temporal consistency following the reference video.
    The reference videos are shown in the lower left corner of the generated results, respectively.
    \textcolor{black}{The input images are synthetic images produced by publicly available LoRA models~\cite{civitai}.}
    }
    \label{fig:teaser}
\end{figure*}

Recent advances in video diffusion models~\cite{team2025kling, yang2025unified, yang2024cogvideox, agarwal2025cosmos} have substantially enhanced digital content creation by enabling high-quality visual effects (VFX) synthesis~\cite{li2025vfxmaster}. Such progress has facilitated realistic and efficient VFX generation across applications, including film production, augmented reality, and interactive gaming, while promising to significantly reduce the cost and effort required for professional-grade content. However, despite these capabilities, deploying generalist models for specialized VFX tasks remains challenging, particularly when operating under strictly limited computational resources.

Current research in VFX generation follows two paradigms generally, yet both struggle to balance efficiency with generalization. The first paradigm relies on per-effect optimization—such as LoRA-based fine-tuning~\cite{hu2022lora, liu2025vfx}—which is computationally efficient but suffers from limited flexibility. Since these methods require a dedicated training cycle for every new effect, they fail to generalize to out-of-distribution scenarios. Conversely, the second paradigm focuses on large-scale supervised training on massive datasets~\cite{li2025vfxmaster, bian2025video}, leveraging in-context learning to capture broad effect categories. While these models offer better generalization, they are resource-intensive, often requiring massive clusters and extensive high-quality data, as shown in Fig.~\ref{fig:efficiency}. Despite these heavy requirements, such models still exhibit performance degradation when encountering specialized effects that involve large distributional shifts from the training data.

\begin{figure}[t]
    \centering
    \includegraphics[width=0.9\linewidth]{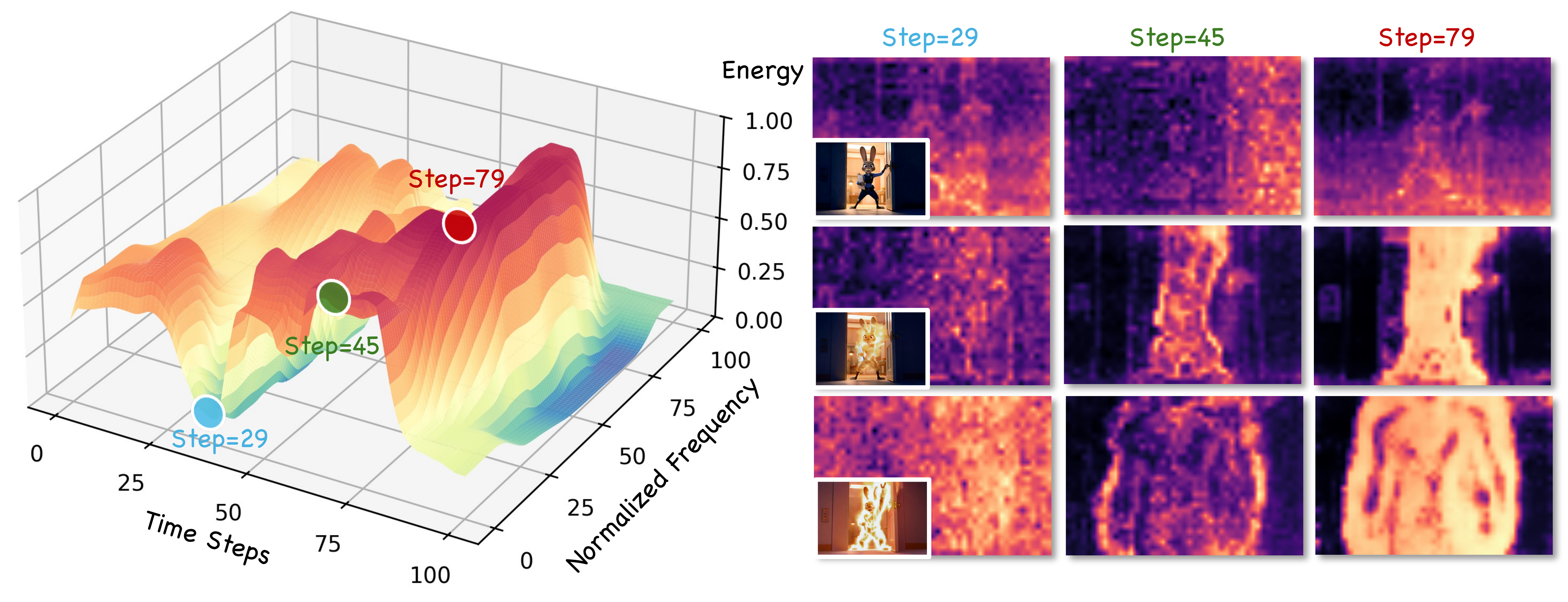}
    \caption{\textbf{Analysis of spectral frequency components during visual effect generation.}
The left part plots the overall spectral energy evolution over diffusion time steps across normalized frequencies. The right panels visualize intermediate spatial features at select steps ($t=29, 45, 79$), demonstrating that high-frequency components (bottom rows) consistently encode fine-grained visual effect details, while low-frequency components (top rows) primarily capture global structure. This distinct, complementary behavior strongly motivates our frequency-decoupled approach.}
    \label{fig:motivation}
\end{figure}

\begin{figure}[t]
    \centering
    \includegraphics[width=0.8\linewidth]{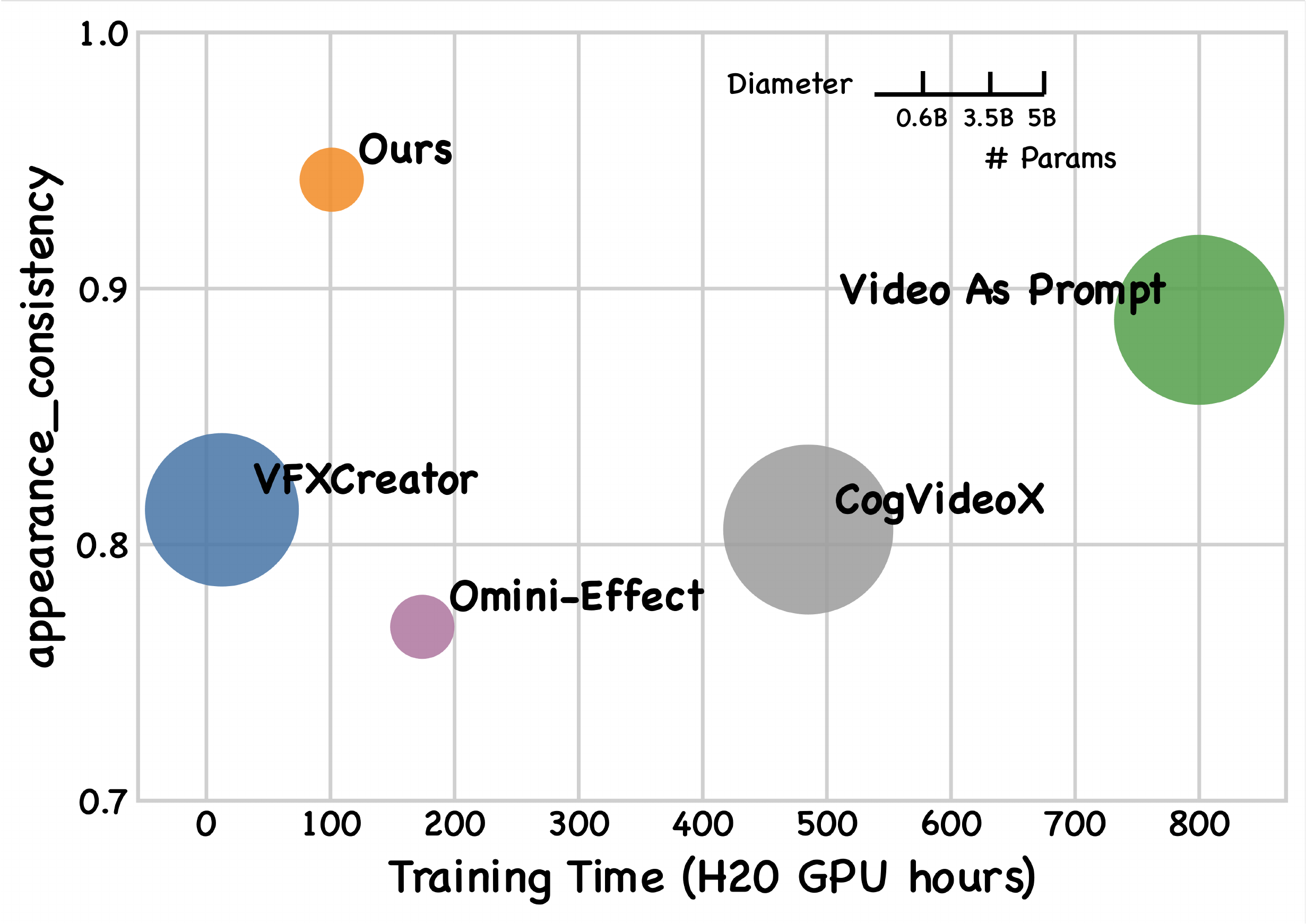}
    \caption{\textbf{Comparison of training efficiency and generation quality.}
We compare EasyVFX with video generation and VFX baselines under the experimental settings described in Sec.~\ref{sec:experiments}. All CogVideoX-based methods are initialized from the same pretrained CogVideoX checkpoint, and \textcolor{black}{the reported CogVideoX cost refers to fine-tuning on OpenVFX rather than full training from scratch.} EasyVFX achieves a favorable trade-off between appearance consistency and training/adaptation cost by keeping the pretrained backbone frozen and optimizing only lightweight VFX-related modules.}
    \label{fig:efficiency}
\end{figure}

Our objective is to develop a framework capable of producing high-fidelity VFX while remaining trainable under stringent resource constraints, which enables standard laboratories to achieve professional-grade results without massive compute. This motivates us to exploit the intrinsic characteristics of VFX synthesis within diffusion models. Unlike traditional style transfer, which primarily involves texture mapping, VFX generation often requires a simultaneous and complex transformation of both texture details and global coherent dynamics. We empirically  observe that VFX signals exhibit distinct signatures across spatial frequency bands and temporal variation cues in the frequency domain. As illustrated in Fig.~\ref{fig:motivation}, the spectral energy evolves noticeably across diffusion time steps, where feature visualizations at $t=29, 45, 79$ reveal a clear division of roles: high-frequency components preserve fine-grained details—such as the sharp edges of lightning or granular smoke textures—defining the effect's motion details, while low-frequency components capture global structure and coherent dynamics.

However, existing paradigms fail to explicitly leverage this spectral distribution. Due to the inherent spectral bias of diffusion models, the optimization process naturally gravitates toward low-frequency signals, leaving delicate high-frequency textures insufficiently learned. Whether employing massive-scale pre-training or per-effect LoRA fine-tuning, current methods treat these diverse components uniformly, essentially "brute-forcing" the learning process without an explicit understanding of frequency-dependent information. This frequency-agnostic approach is inherently inefficient and resource-intensive, often sacrificing textural realism for motion stability when compute is limited. We hypothesize that by enabling the diffusion process to explicitly learn frequency-specific features, we can achieve superior VFX quality and robust generalization with significantly fewer computational resources.

In this paper, we propose \textbf{EasyVFX}, a two-stage framework that leverages spectral cues for efficient visual effect generation. In Stage 1, we introduce a Frequency-aware Mixture-of-Experts (Freq-MoE) architecture with frequency routers, where experts are encouraged to specialize in different coarse spectral characteristics of visual effects. Crucially, this stage is trained only once with a lightweight schedule, distilling general VFX priors that capture common effect structure and dynamics. In Stage 2, to bridge the gap between these shared priors and reference-specific appearance, we adopt a Test-Time Training (TTT) strategy. \textcolor{black}{Given a single reference, we optimize a compact VFX embedding with a frequency-constraint loss, which serves as a lightweight spectral regularization term by encouraging consistency between normalized frequency-energy statistics of the generated and reference videos. This design separates “learn once” generalization from “adapt fast” specialization and provides a favorable quality--efficiency trade-off.}

\begin{itemize}
    \item We analyze the spectral trait of visual effects videos, revealing distinct low/high-frequency roles, and propose \textbf{EasyVFX}, a frequency-aware framework for VFX generation.
    \item Technically, we design a two-stage strategy: (i) a \textbf{Frequency-aware Mixture-of-Experts (Freq-MoE)} to encourage frequency-aware expert specialization, and (ii) a \textbf{Test-Time Training (TTT)} strategy with an effect embedding and \textcolor{black}{a frequency-constraint loss that acts as a lightweight spectral regularization term for fast instance-level adaptation.}
    \item Extensive quantitative and qualitative experiments demonstrate that EasyVFX achieves superior visual quality with high resource-efficiency, significantly outperforming state-of-the-art baselines in both objective metrics and human preference.
\end{itemize}


\section{Related work}

\noindent \textbf{Controllable Video Generation.}
With the advancement of video generation models, large-scale parameter-based video generation models~\cite{brooks2024video, kong2025hunyuanvideosystematicframeworklarge, ma2026group, xu2025clgc, xu2025smrabooth, hu2026embeddingperturbed, qiu2024tfb, qiu2025duet, qiu2025DBLoss, qiu2026dag, ma2024followpose, ma2025followcreation, ma2026fastvmt, ma2025followyourmotion, ma2025controllable, ma2025followfaster} have shown strong generation effects. These models can generate corresponding video content based on users' text prompts by learning a large amount of text-video paired data. However, the form of text-to-video generation relying solely on text-based conditions is difficult to convey complex spatiotemporal semantics accurately. To address this issue, controllable video generation technology ~\cite{hu2024animate, zhang2025easycontrol,xue2025infinihuman,xue2026georelight, zhang2024ssr, song2025layertracer, ma2022visual, song2025makeanything, wang2024taming, feng2025dit4edit, yan2025eedit, yu2026latent, liu2025avatarartist, ma2024followyouremoji, ma2025followyourclick, guo2024liveportrait, jiang2025vace, zhang2025flexiact} has emerged in recent years, aiming to generate videos accurately based on user intent, producing highly customized content. Controllable video generation not only relies on text descriptions as the sole input condition, but also introduces other modal information, such as visual~\cite{wang2025frame, fei2025skyreels,zhu2024champ}, spatial~\cite{wang2019point, xue2024human, xu2024magicanimate, long2025follow, shen2025follow, wang2025unianimate}, and temporal~\cite{lei2025ditraj, he2024cameractrl, wu2024draganything}. By incorporating these additional modalities, users are able to have finer-grained control over the generation process. For instance, using images as starting frames~\cite{Zhang_2025_CVPR, 10.1145/3581783.3611897} or reference subjects~\cite{DreamVideo, CustomVideo, Huang_2025_CVPR} can help the model determine the overall style and structure of the video, while depth maps~\cite{zhou2025holotime, guo2024sparsectrl, xing2024make, peng2024controlnext} and motion trajectories~\cite{geng2025motion, yariv2025through, ling2024motionclone} guide the model to generate dynamic effects that conform to physical laws. This control mechanism enables the model to generate efficiently in more diverse and complex application scenarios, overcoming the limitations brought by using text descriptions alone.

\noindent \textbf{Visual Effects Generation.}
Visual Effects (VFX) Generation has seen significant advancements in recent years, providing an efficient alternative to traditional production methods that rely on procedural methods and artist expertise. However, despite progress in general video generation, controllable VFX generation remains an underexplored area due to the lack of diverse VFX data and limitations in conditional control. Early works, like MagicVFX~\cite{guo2024magicvfx}, were limited to simple overlay effects, lacking flexibility and scalability. VFXCreator~\cite{liu2025vfx} introduced controllable VFX by training separate LoRA modules for each effect, but it is restricted to generating single effects, limiting its adaptability. Omni-Effects~\cite{mao2025omni} introduced a more generalized solution by employing LoRA-MoE, enabling the generation of multiple composite effects within a unified framework. Despite this, it remains confined to in-domain combinations and struggles with generalizing to novel, out-of-domain effects.
While these advancements mark a shift from single-effect models to more flexible approaches, they still face challenges in adapting to diverse video contexts. VFXMaster~\cite{li2025vfxmaster} employs a referential learning approach to transfer visual effects from reference videos to target videos, providing a more general solution capable of handling the generation of effects in different video content. However, this method still faces certain limitations, particularly in generating diverse and complex open-domain effects, where it cannot flexibly adapt to dynamic changes in different scenes.

\begin{figure*}[t]
    \centering
    \includegraphics[width=\linewidth]{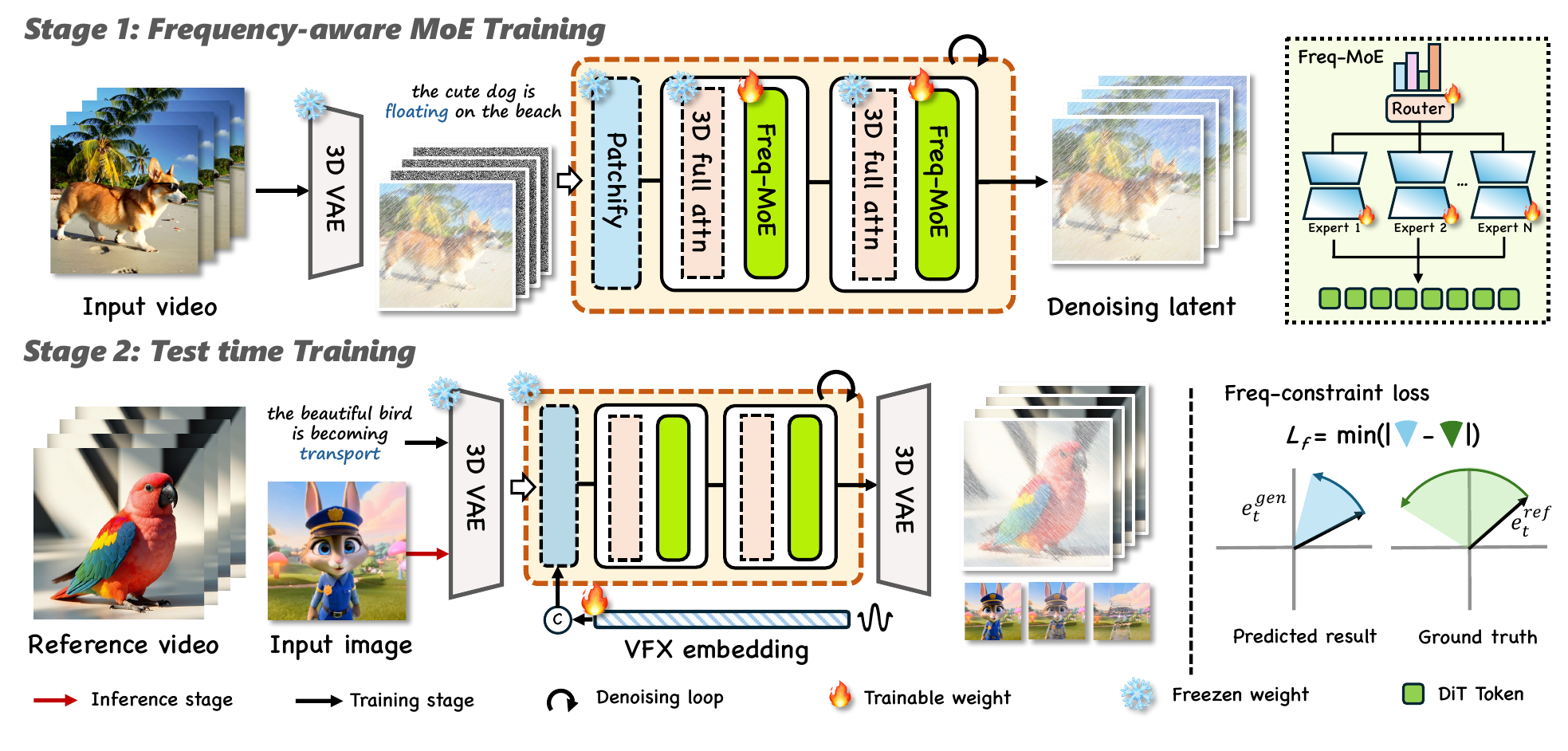}
    \vspace{-15pt}
    \caption{\textbf{Overview of our method.}
    \textcolor{black}{Our method consists of two stages.
    \textbf{\textit{Stage 1: Frequency-aware MoE Training.}}
    We employ a 3D VAE~\cite{kingma2013auto} to project input videos into a latent space. The core architecture uses a Frequency-aware Mixture-of-Experts (Freq-MoE) adapter, where a lightweight router assigns soft weights to LoRA experts according to coarse spectral energy cues extracted from noisy latents. This encourages the adapter to respond differently to effects with different frequency characteristics, such as smooth illumination changes and high-frequency particle effects.
    \textbf{\textit{Stage 2: Test-Time Adaptation.}}
    For reference-specific adaptation, we introduce a small set of learnable VFX embeddings while keeping the diffusion backbone and Stage-1 adapters frozen. The embeddings are optimized with a lightweight spectral regularization term, denoted as $\mathcal{L}_f$, which matches normalized multi-scale frequency-energy descriptors between the generated and reference videos. This regularizer provides coarse spectral guidance rather than enforcing exact spatial or temporal correspondence. Fire and snowflake icons denote trainable and frozen parameters, respectively.}}
    \label{fig:Overview}
\end{figure*}


\section{Method}
\label{sec:method}

\textcolor{black}{Controllable visual effect (VFX) generation involves providing detailed dynamic guidance together with a text prompt, with the goal of producing cinematic VFX videos that follow a reference effect while preserving the target content. In this work, we present \textbf{EasyVFX}, a reference-based framework that extends image-to-video (I2V) diffusion models to reproduce diverse dynamic effects in an in-context manner. Given a single reference VFX video, \textbf{EasyVFX} transfers its overall dynamic effect pattern to a target image while keeping the diffusion backbone frozen during test-time adaptation. Our method introduces a frequency-aware adapter for general VFX priors and a lightweight spectral regularizer for reference-specific adaptation.}

Our framework has two stages. In Sec.~\ref{subsec:prelim}, we introduce the backbone I2V diffusion model and the learning objective. In Sec.~\ref{subsec:stage1}, we present \textbf{Stage~1: Frequency-aware MoE Training}, a \emph{one-time} training stage with a \emph{lightweight} (few steps) schedule that distills general, domain-agnostic VFX priors. In Sec.~\ref{subsec:stage2}, we introduce \textbf{Stage~2: Test-Time Training with VFX Embedding}, which enables fast per-reference adaptation by enforcing spectral alignment between the generated result and the reference. Finally, Sec.~\ref{subsec:workflow} summarizes the overall training and inference workflow, highlighting the separation between both \emph{learn once} generalization and \emph{adapt fast} specialization for efficient high-quality VFX generation.

\subsection{Preliminaries}
\label{subsec:prelim}

\noindent \textbf{Backbone I2V diffusion model.}
We build on a latent diffusion image-to-video model. Given an input image $I\in\mathbb{R}^{H\times W\times 3}$ and a text prompt $p$, the model generates a video
$V\in\mathbb{R}^{T\times H\times W\times 3}$.
A video VAE encodes videos into latents $z\in\mathbb{R}^{B\times T\times C\times h\times w}$, where $(h,w)$ is the latent spatial resolution and $C$ is the latent channel size.
The denoiser is a diffusion transformer conditioned on the input image and the text embedding $\mathbf{g}$ from a frozen text encoder.
We keep the backbone architecture and follow its standard cross-attention conditioning interface; our method also optionally appends additional conditioning tokens (Sec.~\ref{subsubsec:tokens}) to the conditioning sequence.

\noindent \textbf{Objective.}
Let $z_t$ denote the noised latent at diffusion timestep $t$, obtained by the forward diffusion process
$
z_t = \alpha_t z_0 + \sigma_t \epsilon
$
with $\epsilon\sim\mathcal{N}(0,\mathbf{I})$.
The denoising network $\epsilon_{\Theta}(\cdot)$ predicts the diffusion target from $(z_t,t,\mathrm{cond})$, where $\mathrm{cond}$ denotes the frozen conditioning derived from $(I,p)$ (image condition + text tokens).
Training minimizes the mean squared error to the ground-truth target:
\begin{equation}
\label{eq:diff_obj}
\mathcal{L}_{\text{diff}}(\Theta)
=
\mathbb{E}_{z_0,\,t,\,\epsilon}
\left[
\left\|
\epsilon - \epsilon_{\Theta}(z_t, t, \mathrm{cond})
\right\|_2^2
\right].
\end{equation}

\subsection{Frequency-aware MoE Training}
\label{subsec:stage1}

As shown in Fig.~\ref{fig:Overview}, Stage~1 trains a frequency-aware mixture-of-experts (MoE)~\cite{shazeer2017outrageously, zhang2025flexiactflexibleactioncontrol} adapter attached to attention projections for VFX generation. The key idea is to derive an interpretable \emph{frequency-energy indicator} (FEI) from the current noisy latent video and use it as a routing signal to dynamically mix multiple low-rank experts. This is particularly suitable for VFX because different effects exhibit distinct spectral signatures (e.g., smooth low-frequency illumination versus high-frequency particles/sparks), and such signatures evolve throughout the denoising trajectory, providing a stable visual effect cue.

\noindent \textbf{\textit{Frequency-energy indicator (FEI).}}
Following the energy-based routing intuition in FeRA~\cite{FeRA}, we derive a compact frequency-energy indicator at each diffusion step $t$. Given the noisy latent video $z_t\in\mathbb{R}^{B\times T\times C\times h\times w}$, we first apply a fixed temporal reducer $\mathcal{A}(\cdot)$ (Eqs.~\eqref{eq:appearance_proxy_stage1}--\eqref{eq:motion_proxy_stage1}) to obtain a 2D proxy feature
$x_t=\mathcal{A}(z_t)\in\mathbb{R}^{B\times C\times h\times w}$.
We then apply a simple two-scale spatial low-pass operator $\mathcal{L}_{\sigma}(\cdot)$ (implemented as a depthwise Gaussian smoothing in our code) to form three coarse-to-fine components:
a coarse component $\mathcal{L}_{\sigma_2}(x_t)$, a band-pass component $\mathcal{L}_{\sigma_1}(x_t)-\mathcal{L}_{\sigma_2}(x_t)$, and a detail residual $x_t-\mathcal{L}_{\sigma_1}(x_t)$, with $\sigma_1<\sigma_2$.
For each component $k\in\{0,1,2\}$, we compute a per-sample energy $E_{t,k}\in\mathbb{R}^{B}$ by summing squared activations over the channel and spatial dimensions of the corresponding component, yielding scale-wise energy cues.

We normalize $\{E_{t,k}\}$ to obtain a relative energy distribution:
\begin{equation}
\label{eq:energy_stage1}
\mathbf{e}_t = \frac{[E_{t,0},E_{t,1},E_{t,2}]}{E_{t,0}+E_{t,1}+E_{t,2}+\varepsilon}\in\mathbb{R}^{B\times 3},
\end{equation}
where $\varepsilon$ is a small constant for numerical stability.

\noindent \textit{\textbf{Dual-component signals.}}
\label{subsubsec:dual}
For noisy latent video $z_t \in \mathbb{R}^{B\times T\times C\times h\times w}$ at diffusion step $t$, we instantiate the temporal reducer $\mathcal{A}(\cdot)$ as two fixed aggregation operators and compute FEI on each resulting 2D proxy. Concretely, the appearance reducer $\mathcal{A}_{\text{app}}$ averages features over time to obtain $x^{\text{app}}_t\in\mathbb{R}^{B\times C\times h\times w}$, capturing temporally stable content, whereas the VFX reducer $\mathcal{A}_{\text{VFX}}$ aggregates frame-to-frame variations to obtain $x^{\text{VFX}}_t\in\mathbb{R}^{B\times C\times h\times w}$, reflecting the visual effects intensity. These two complementary proxies provide informative cues for expert selection across different noise levels.

We aggregate over time to emphasize temporal consistency:
\begin{equation}
\label{eq:appearance_proxy_stage1}
x^{\text{app}}_t
= \mathcal{A}_{\text{app}}(z_t)
= \frac{1}{T}\sum_{i=1}^{T} z_{t,i}
\in\mathbb{R}^{B\times C\times h\times w}.
\end{equation}

We highlight temporal variation via frame differences:
\begin{gather}
\label{eq:motion_proxy_stage1}
D_{t,i} = z_{t,i+1} - z_{t,i}\\
x^{\text{VFX}}_t
= \mathcal{A}_{\text{VFX}}(z_t)
= \log\!\left(1 + \frac{1}{T-1}\sum_{i=1}^{T-1} D_{t,i}^{2}\right)
\in\mathbb{R}^{B\times C\times h\times w},
\end{gather}
where the square is applied element-wise. $\log(1+\cdot)$ compression reduces the dynamic range of VFX magnitudes and stabilizes routing.

We then compute $\mathbf{e}^{\text{app}}_t=\mathrm{FEI}(x^{\text{app}}_t)$ and $\mathbf{e}^{\text{VFX}}_t=\mathrm{FEI}(x^{\text{VFX}}_t)$, both in $\mathbb{R}^{B\times 3}$, and concatenate them into a joint routing descriptor:
\begin{equation}
\label{eq:joint_descriptor_stage1}
\mathbf{e}_t
=
\mathrm{FEI}_{\text{joint}}(z_t)
=[\mathbf{e}^{\text{app}}_t,\mathbf{e}^{\text{VFX}}_t]\in\mathbb{R}^{B\times 6}.
\end{equation}
This dual-component signal helps decouple what should remain appearance-consistent from what should evolve over time.

\noindent \textit{\textbf{MoE LoRA experts and frequency router.}}
\label{subsubsec:router}
Given the joint frequency-energy descriptor $\mathbf{e}_t$ at diffusion step $t$, we route computation to $M$ experts with a lightweight router. A two-layer MLP $r_{\phi}$ maps $\mathbf{e}_t$ (applied per sample) to $M$ routing logits, which are converted to mixture weights by a temperature-controlled softmax. We use soft routing to keep optimization stable across diverse VFX categories and to support smooth interpolation when an effect exhibits mixed spectral characteristics. In our implementation, $\mathbf{e}_t$ is computed from the current noisy latent and detached from gradients.

We attach $M$ LoRA experts to the attention projection layers (e.g., $\mathbf{W}_q,\mathbf{W}_k,\mathbf{W}_v,\mathbf{W}_o$). For a projection weight $\mathbf{W}$ and hidden state $\mathbf{h}$, expert $m$ produces a low-rank additive update $\Delta\mathbf{W}^{(m)}\mathbf{h} = \mathbf{B}^{(m)}\mathbf{A}^{(m)}\mathbf{h}$ (down-projection $\mathbf{A}^{(m)}$, up-projection $\mathbf{B}^{(m)}$). The router mixes experts by a weighted sum of their updates and adds the result to the projection output with a standard LoRA scaling factor $s$:
\begin{equation}
\Delta\mathbf{h} = s\sum_{m=1}^{M} \pi_{t}^{(m)} \, \mathbf{B}^{(m)}\mathbf{A}^{(m)}\mathbf{h},
\end{equation}
where $\pi_{t}^{(m)}$ are the softmax mixture weights from $r_{\phi}(\mathbf{e}_t)$. To keep the parameter budget comparable, we fix a total rank $r$ and distribute it across experts so that $\sum_{m=1}^{M} r_m = r$.
In Stage~1, we update only the router parameters $\phi$ and the expert adapters.

\subsection{Test-Time Training}
\label{subsec:stage2}

Stage~2 adapts VFX from a reference video to a target generation by updating only a small set of \emph{VFX embeddings}, while freezing the base I2V diffusion backbone, the frequency router, and all LoRA experts. Concretely, given a reference video exhibiting the desired VFX pattern (e.g., particle flow, flame flicker, or camera shake), we perform lightweight test-time optimization that adjusts only the conditioning tokens, so that the VFX spectrum of the generated video matches the reference video for better consistency.

\noindent \textit{\textbf{VFX embedding.}}
\label{subsubsec:tokens}
We introduce a learnable embedding set $\mathbf{V}\in\mathbb{R}^{L\times d}$ with $L$ embeddings of dimension $d$. During inference, $\mathbf{V}$ is broadcast to the batch dimension and concatenated to the text encoder hidden states, serving as additional conditioning tokens in cross-attention.

\noindent \textit{\textbf{Freq-constraint loss.}}
\label{subsubsec:freq_loss}
Let $z^{\text{ref}}_t, z^{\text{gen}}_t \in \mathbb{R}^{B\times T\times C\times h\times w}$ denote the noisy latent videos of the reference and the current generation at diffusion step $t$. Using the same dual-component temporal reducers in Sec.~\ref{subsubsec:dual}, we can define the test-time frequency constraint as the L1 distance between the two normalized energy-ratio descriptors:
\begin{equation}
\label{eq:Lf}
\mathcal{L}_{f}(t)
=
\left\lVert
\mathbf{e}^{\text{gen}}_{t} -
\mathbf{e}^{\text{ref}}_{t}
\right\rVert_{1}.
\end{equation}
Since each descriptor is normalized per sample, $\mathcal{L}_f$ is robust to absolute scale differences and directly aligns the low/mid/high energy distribution of both appearance and VFX components, yielding a stable objective for adaptation under diffusion noise.

\noindent{\textbf{Stage~2 objective.}}
\label{para:stage2_obj}
In Stage~2, we optimize only the VFX embeddings $\mathbf{V}$ by minimizing the expected frequency constraint loss:
\begin{equation}
\label{eq:stage2_obj}
\min_{\mathbf{V}} \ \mathbb{E}_{t\sim\mathcal{U}(\mathcal{S})}\left[\mathcal{L}_f(t)\right],
\end{equation}
where $\mathcal{S}$ denotes a chosen set of timesteps for test-time training (e.g., mid-to-late denoising steps), optionally, Eq.~\eqref{eq:stage2_obj} can be combined with a standard denoising objective; in our setting, the frequency matching loss alone provides efficient adaptation and adds negligible test-time overhead while keeping all other parameters frozen.

\subsection{Training and Inference Workflow}
\label{subsec:workflow}
Fig.~\ref{fig:Overview} summarizes our workflow. The pipeline consists of an offline training stage and a lightweight test-time adaptation stage.

\paragraph{Stage~1 (offline).}
Stage~1 learns a general VFX adapter on top of a frozen I2V diffusion backbone. We attach a frequency-aware MoE LoRA adapter to attention projection layers. During training, we compute the FEI routing signal from the current noisy latent using the appearance and VFX proxies, and use a soft router to mix multiple LoRA experts. We optimize only the router and expert parameters with the standard diffusion objective in Eq.~\eqref{eq:diff_obj}.

\paragraph{Stage~2 (test-time).}
Stage~2 adapts to a specific reference VFX video without updating the backbone or the MoE adapter. We append a small set of learnable token embeddings $\mathbf{V}$ to the conditioning token sequence for cross-attention. Given a reference video, we update only $\mathbf{V}$ by matching the dual-component FEI statistics of the generated noisy latents to those computed from the reference for better consistency (Eq.~\eqref{eq:stage2_obj}). This yields one-shot, reference-specific adaptation with minimal computation at test time.

\paragraph{Inference.}
After Stage~2, we fix the adapted tokens and run standard I2V sampling conditioned on the target image and prompt. The frequency-aware router remains active to mix experts along the denoising trajectory, while the adapted tokens steer the generation toward the reference effect dynamics more faithfully.

\begin{figure*}[t]
    \centering
    \includegraphics[width=0.8\linewidth]{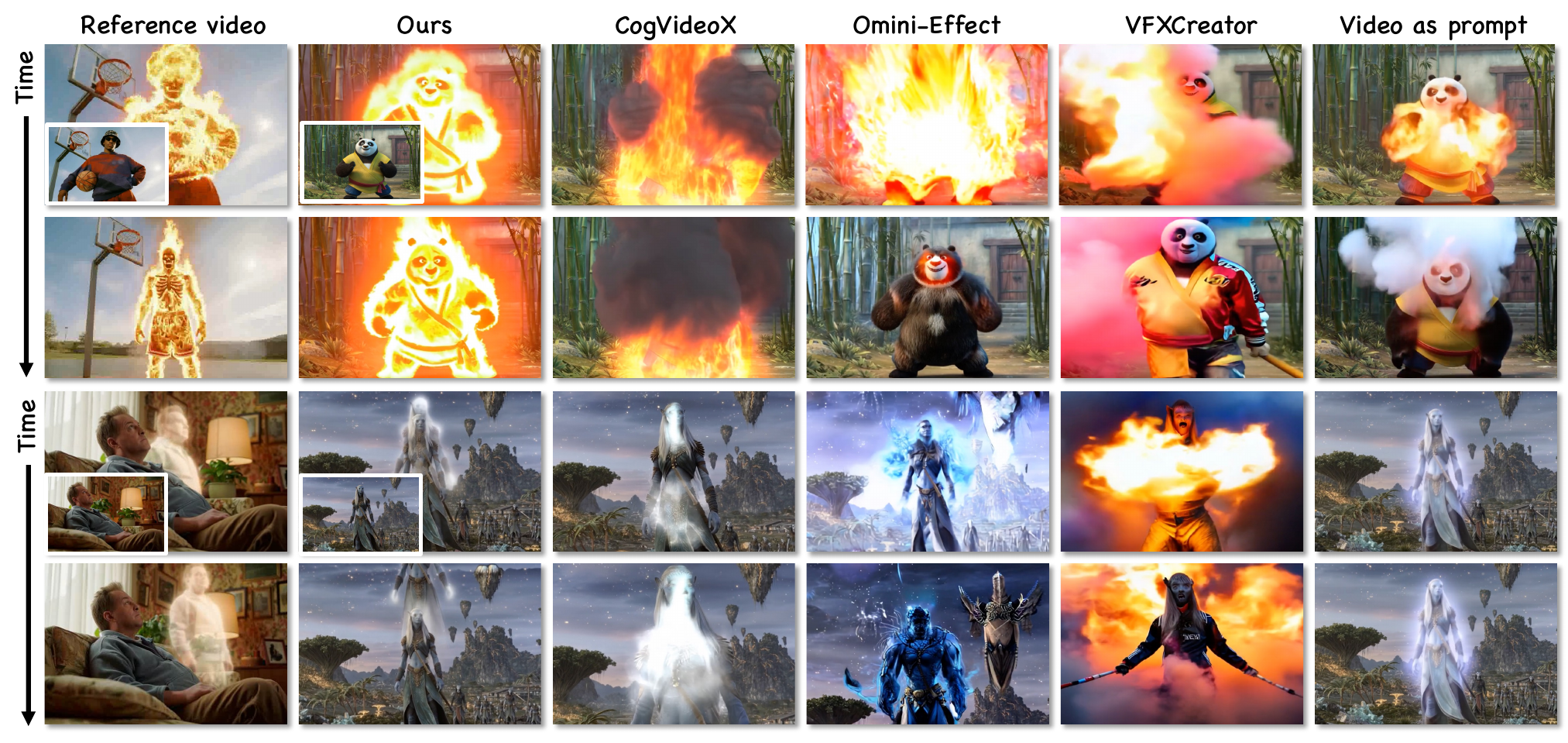}
    \caption{\textcolor{black}{\textbf{Qualitative comparison results with existing methods.} EasyVFX produces visually plausible effects and coherent temporal evolution by leveraging frequency-aware expert routing to capture coarse effect appearance and motion cues.}}
    \label{fig:comparison}
\end{figure*}

\begin{table*}[t]
  \centering
  \caption{\textbf{Comparison with state-of-the-art visual effect generation methods}. 
  For fair comparison, we select CogVideoX~\cite{yang2024cogvideox} as the same backbone. 
  \textcolor{Red}{\textbf{Red}} and \textcolor{Blue}{\textbf{Blue}} denote the best and second best results, respectively.}
  \label{tab:comparison}
  \resizebox{0.9\textwidth}{!}{%
    \begin{tabular}{l|cccc|cccc}
      \toprule
      \multirow{2}{*}{Method}
        & \multicolumn{4}{c|}{Quantitative Metrics}
        & \multicolumn{4}{c}{VBench Metrics} \\
      \cmidrule(lr){2-5}\cmidrule(lr){6-9}
        & Text Sim.$\uparrow$ & Motion Fid.$\uparrow$ & Temp. Cons.$\uparrow$ & App. Cons.$\uparrow$
        & Sub. Cons.$\uparrow$ & Back. Cons.$\uparrow$ & Aes. Qual.$\uparrow$ & Motion Smooth.$\uparrow$ \\
      \midrule
      VFX Creator~\citep{liu2025vfx}
        & 0.2218
        & \textcolor{Blue}{\textbf{0.3761}}
        & 0.8718
        & 0.8136
        & 0.8215 & 0.8833 & 0.5127 & \textcolor{Blue}{\textbf{0.9827}} \\
      Video-As-Prompt~\citep{bian2025video}
        & \textcolor{Blue}{\textbf{0.3051}}
        & 0.3693
        & \textcolor{Blue}{\textbf{0.8940}}
        & \textcolor{Blue}{\textbf{0.8878}}
        & \textcolor{Blue}{\textbf{0.8441}} & \textcolor{Blue}{\textbf{0.9023}} & \textcolor{Blue}{\textbf{0.6066}} & 0.9838\\
      Omini-Effects~\citep{mao2025omni}
        & 0.2156 & 0.3568 & 0.8278 & 0.7679
        & 0.7642 & 0.8540 & 0.4679 & 0.9690 \\
      CogVideoX~\citep{yang2024cogvideox}
        & 0.2724
        & 0.3720
        & 0.8790
        & 0.8059
        & 0.7949 & 0.8801 & 0.5809 & 0.9718 \\
      Ours
        & \textcolor{Red}{\textbf{0.3274}}
        & \textcolor{Red}{\textbf{0.3985}}
        & \textcolor{Red}{\textbf{0.9522}}
        & \textcolor{Red}{\textbf{0.9425}}
        & \textcolor{Red}{\textbf{0.9077}} & \textcolor{Red}{\textbf{0.9210}} & \textcolor{Red}{\textbf{0.6542}} & \textcolor{Red}{\textbf{0.9886}}  \\
      \bottomrule
    \end{tabular}%
  }
    \vspace{-10pt}
\end{table*}

\section{Experiments}

\subsection{Implementation details}
In our experiments, we train our model using CogVideoX-5B~\cite{yang2024cogvideox}. During frequency-aware MoE training, we perform a \textbf{one-time} 5000-step training on OpenVFX~\cite{liu2025vfx} using 8 NVIDIA H20. The optimization is performed using AdamW~\cite{loshchilov2017decoupled} with a weight decay of 0.01 and an initial learning rate of $1\times10^{-5}$. 
For the test time training stage,  EasyVFX requires 50 to
100 training steps on each reference video using 1 NVIDIA H20, depending on visual effect complexity. In comparison, the VFX-Creator needs 3,000 steps
for specific LoRA. \textcolor{black}{Note that we only used VAP-Data~\cite{bian2025video} in a few cases for visualization and qualitative demonstration purposes}. More  details are provided in the supplementary materials.



\subsection{Comparison with baselines}
\noindent \textbf{Qualitative results.}
We compare our approach with the state-of-the-art visual effect generation methods visually. They include the feed-forward methods, such as Video-as-Prompt~\citep{bian2025video}, Omini-Effects~\citep{mao2025omni}, CogVideoX~\citep{yang2024cogvideox}, and LoRA-based methods, VFX Creator~\citep{liu2025vfx}. \textcolor{black}{Note that VFX Creator and EasyVFX use reference-specific test-time adaptation. }
For fair comparison, we adapt the CogVideoX-5B~\citep{yang2024cogvideox} as the same backbone. 
Our experimental results demonstrate that EasyVFX achieves superior performance and greater versatility across a wide range of in-domain and out-of-domain visual effect scenarios.
As illustrated in Fig.~\ref{fig:comparison}, these works~\citep{mao2025omni, yang2024cogvideox, liu2025vfx}
have the challenge of handling complicated visual effects.  In contrast, our method enables generating videos with aligned effect patterns, preserving the visual consistency with reference videos. We provide more results in the supplementary materials.


\noindent \textbf{Quantitative results.}
We compare our method with the 
state-of-the-art visual effect generation
on 200 high-quality videos randomly selected from the OpenVFX test set~\cite{liu2025vfx}. 
For fair comparison, we employ CogVideoX~\citep{yang2024cogvideox} as the same backbone. All methods are evaluated on the same set of (prompt, target image, reference effects video) inputs. \textcolor{black}{We select four metrics for evaluation. We do not use the VFX-Cons.~\citep{li2025vfxmaster} as the main metric because it aggregates multiple aspects into a single score and does not explicitly isolate reference-effect faithfulness:}
(a) \textbf{Text Similarity: } CLIP is used to extract features from the target video, and the average cosine similarity between the desired visual effect description and video frames is computed. \textcolor{black}{For OpenVFX, we use the desired visual effect description provided in the prompt. Therefore, Text Similarity measures whether the generated video reflects the requested effect category or effect description.}
(b) \textbf{Motion Fidelity:} We use motion fidelity to assess the similarity of tracklets between reference and generated videos.
(c) \textbf{Temporal Consistency:} We measure frame-to-frame coherence by calculating the average feature similarity of consecutive video frames using CLIP~\citep{radford2021learning}.
(d) \textbf{Appearance Consistency}:  We measure how well the generated video preserves the target effect appearance with respect to the reference video.
The results are reported in Tab.~\ref{tab:comparison}. EasyVFX achieves a strong overall performance across the evaluated quantitative metrics under our experimental setting.

In addition, we randomly collect 100 high-quality generated videos from Omini-Effects~\citep{mao2025omni}.    
Four metrics in VBench~\citep{huang2024vbench} are employed for a more accurate evaluation (in Tab.~\ref{tab:comparison}). 
(1) \textbf{Subject Consistency:} We assess whether the identity of the subject is preserved across frames. 
(2) \textbf{Motion Smoothness:} The metric evaluates inter-frame continuity using learned motion priors. 
(3) \textbf{Aesthetic Quality} uses a LAION-trained aesthetic predictor to score visual appeal.
(4) \textbf{Background Consistency:} We evaluate the coherence of the background.
Our proposed method significantly outperforms all baseline approaches across every video quality metric, thereby showcasing the state-of-the-art performance.

\begin{figure}[t]
    \centering
    \includegraphics[width=0.85\linewidth]{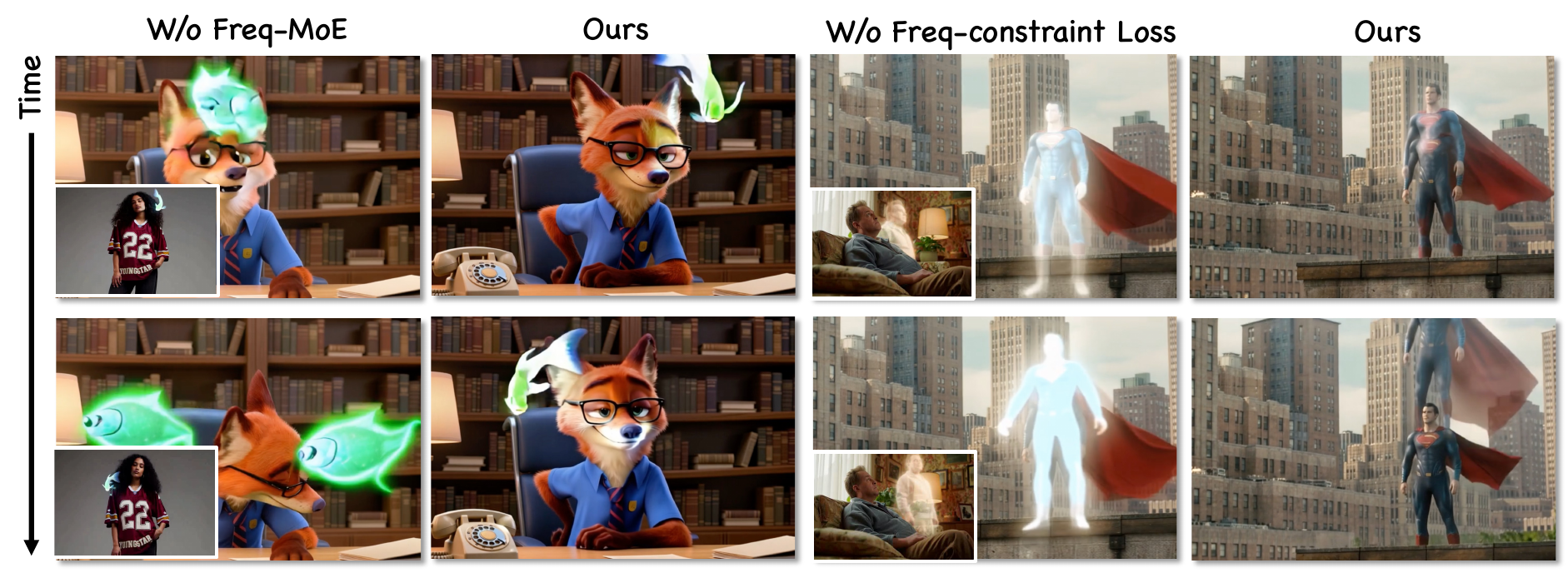}
    \caption{\textbf{Visual ablation of proposed Freq-MoE and Freq-constraint Loss}. We compare the results without the Freq-MoE module (left) and without the Frequency-constraint Loss (right). The ablation study demonstrates that Freq-MoE is essential for better performance, while the Freq-constraint Loss plays a critical role in preserving high-frequency textures.}
    \label{fig:ab_about_model1}
\end{figure}

\begin{figure}[t]
    \centering
    \includegraphics[width=0.85\linewidth]{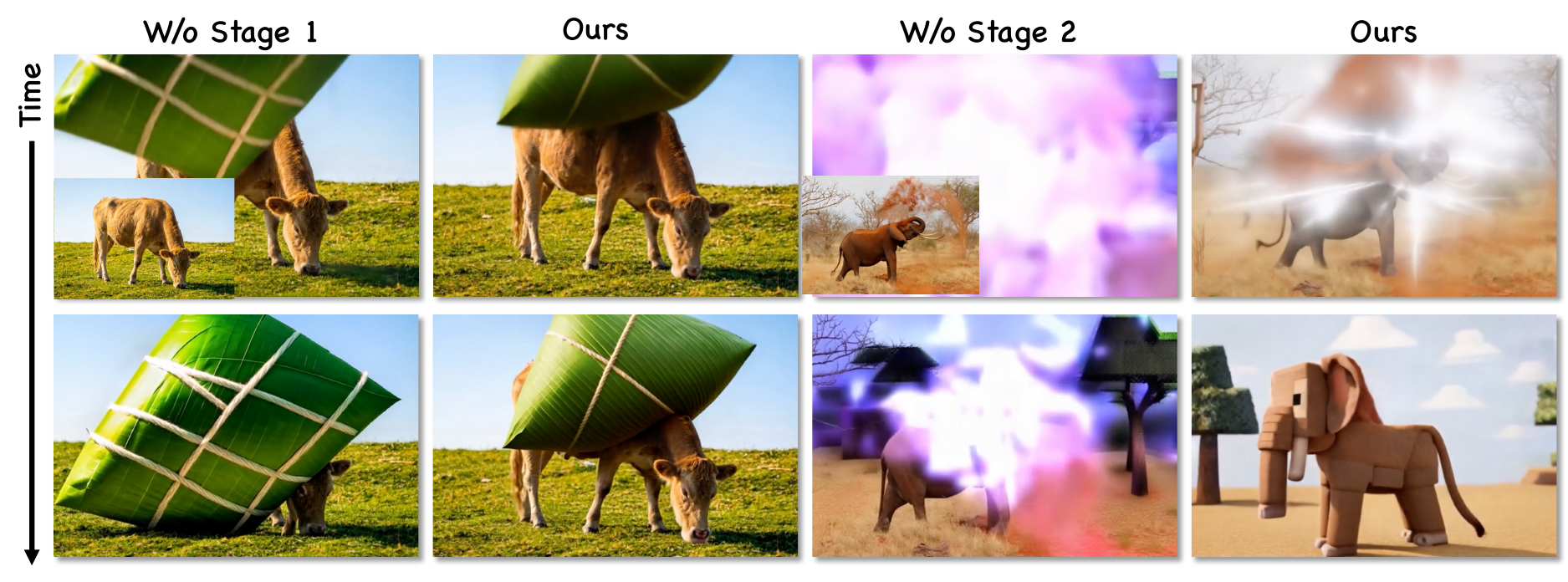}
    \caption{\textbf{Visual ablation of proposed two-stage training.} The results demonstrate that Stage 1 is crucial for visual effect alignment, while Stage 2 is indispensable for recovering high-fidelity details.}
    \label{fig:ab_two_stage}
\end{figure}

 

\subsection{Ablation study}
\textcolor{black}{We provide the ablation study in Fig.~\ref{fig:ab_about_model1}. Without the Freq-MoE module or the frequency-constraint loss, the model tends to produce less stable effect patterns and inconsistent temporal evolution, which can manifest as positional drift or misalignment of the effect regions.}

\begin{table}[t]
  \centering
  \caption{\textbf{Quantitative ablation of proposed Freq-MoE and Freq-constraint Loss}. 
  \textcolor{Red}{\textbf{Red}} and \textcolor{Blue}{\textbf{Blue}} denote the best and second best results, respectively.}
  \label{tab:ab_comparison1}
  \resizebox{0.9\linewidth}{!}{%
    \begin{tabular}{l|cccc}
      \toprule
      \multirow{2}{*}{Method}
        & \multicolumn{4}{c}{Quantitative Metrics} \\
      \cmidrule(lr){2-5}
        & Text Sim.$\uparrow$ & Motion Fid.$\uparrow$ & Temp. Cons.$\uparrow$ & App. Cons.$\uparrow$ \\
      \midrule
      W/o Freq-MoE
        & 0.2788 
        & 0.3217 
        & \textcolor{Blue}{\textbf{0.9316}} 
        & 0.9261 \\
      W/o Freq-constraint Loss
        & \textcolor{Blue}{\textbf{0.2936}} 
        & 0.3564 
        & 0.9284 
        & 0.9317 \\
      \midrule
      W/o Stage 1
        & 0.2764 
        & \textcolor{Blue}{\textbf{0.3715}} 
        & 0.9315 
        & \textcolor{Blue}{\textbf{0.9384}} \\
      W/o Stage 2
        & 0.2619 
        & 0.3623 
        & 0.9118 
        & 0.9253 \\
      \midrule
      Ours
        & \textcolor{Red}{\textbf{0.3274}}
        & \textcolor{Red}{\textbf{0.3985}}
        & \textcolor{Red}{\textbf{0.9522}}
        & \textcolor{Red}{\textbf{0.9425}} \\
      \bottomrule
    \end{tabular}%
  }
\end{table}

\noindent \textbf{Effectiveness of Freq-MoE.}
We conduct quantitative ablation studies to evaluate the effectiveness of the proposed Freq-MoE module under the same experimental settings.
As shown in Tab.~\ref{tab:ab_comparison1} and Fig.~\ref{fig:ab_about_model1}, removing Freq-MoE consistently leads to performance degradation across all quantitative metrics, indicating weakened motion modeling capability. 
In contrast, the full model achieves the best overall performance on all metrics. These results demonstrate that the proposed Freq-MoE plays a critical role in our framework.

\noindent \textbf{Effectiveness of Freq-constraint Loss.}
We further evaluate the effectiveness of the proposed Freq-constraint loss in Tab.~\ref{tab:ab_comparison1} and Fig.~\ref{fig:ab_about_model1}.
It can be observed that removing the freq-constraint loss leads to noticeable visual artifacts, especially in textures, over time. We attribute this degradation to the lack of explicit frequency-level regularization, which results in less stable motion patterns and inconsistent appearance over time. In contrast, incorporating the freq-constraint loss significantly improves temporal coherence and visual stability, validating its effectiveness in our framework.

\noindent \textbf{Effectiveness of two-stage training.}
We evaluate the impact of the proposed two-stage training strategy by comparing it with a single-stage training baseline.
Quantitative results in Tab.~\ref{tab:ab_comparison1} show that abandoning the two-stage design leads to inferior performance, especially in terms of temporal consistency and appearance consistency. As shown in  Fig.~\ref{fig:ab_two_stage}, the two-stage training strategy facilitates more effective learning and results in improved overall generation quality, demonstrating its practical benefit in our method.

\section{Conclusion}

In this paper, we present EasyVFX, a frequency-aware framework for high-fidelity video visual effects generation that is designed to work under limited data and compute. EasyVFX leverages frequency-aware decomposition and expert routing to better capture effect appearance details and motion-related cues, which simplifies optimization and improves generalization. Based on this idea, we propose a Frequency-aware Mixture-of-Experts (Freq-MoE) architecture that assigns specialized experts to different spectral bands. We further introduce a lightweight test-time adaptation strategy with a frequency-based regularization loss (Freq-constraint), enabling fast per-reference adaptation with only a small number of optimization steps. Experiments demonstrate improved visual fidelity and temporal coherence compared to strong baselines, highlighting the practicality of frequency-aware learning for VFX generation.

\tableofcontents
\setcounter{page}{1}
\renewcommand{\thepage}{S\arabic{page}}
\setcounter{section}{0}
\setcounter{figure}{0}
\setcounter{table}{0}

\appendix
\section{Implementary details}
All video inputs are resized to 480p resolution and uniformly sampled to a fixed length of 49 frames. We follow the default preprocessing pipeline of CogVideoX-5B~\cite{yang2024cogvideox} and do not introduce additional modifications to the input representation. For baselines that do not natively support reference-video conditioning, we follow their official setting (no reference conditioning) and still compute Motion Fidelity against the reference for evaluation.
The Frequency-aware Mixture-of-Experts module selects top-$k=3$ experts during routing. This configuration is fixed across all experiments, including both frequency-aware MoE training and test-time adaptation. In Test-time, our VFX Embedding length is set to be 1024. Unless otherwise stated, we run 100 optimization steps per test instance.
During inference, videos are generated using 30 diffusion steps with a classifier-free guidance~\cite{ho2022classifierfreediffusionguidance} scale of 7.5.  All test-time training experiments are conducted on a single NVIDIA RTX A6000 GPU. Only the frequency-aware MoE parameters are updated during test-time adaptation, while the backbone diffusion model remains frozen.

\begin{figure}[t]
    \centering
    \includegraphics[width=\linewidth]{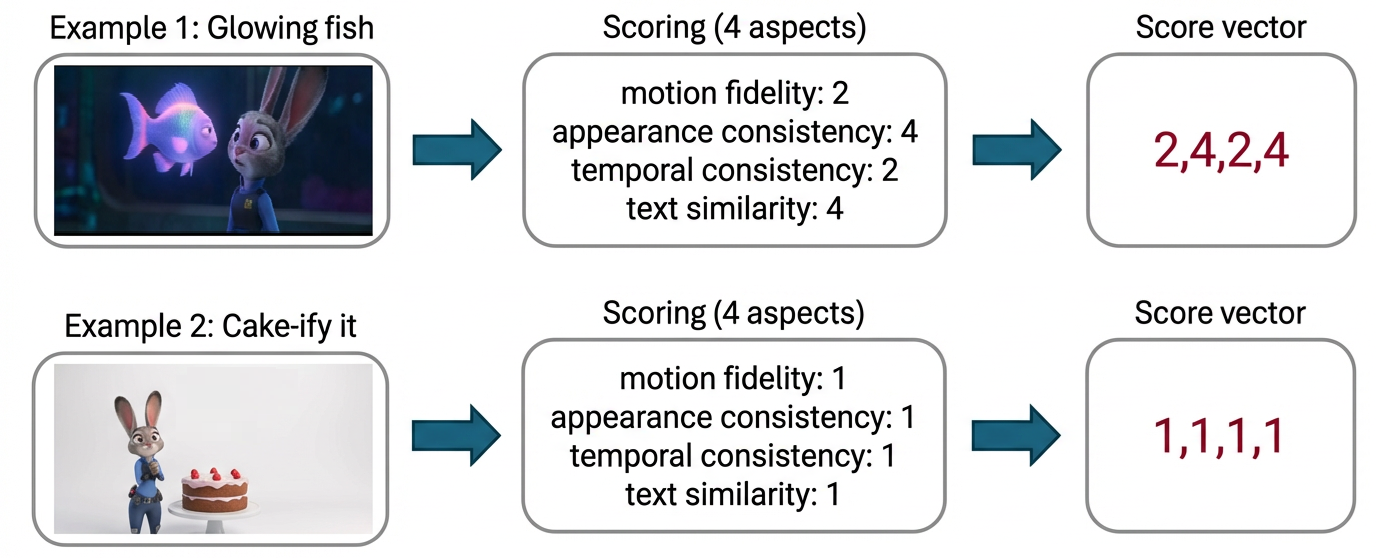}
    \caption{\textbf{User Study Metric}. Given an input prompt and the generated result, raters score (i) motion fidelity, (ii) appearance consistency, (iii) temporal consistency, and (iv) text similarity. We report the four scores as a vector to preserve per-aspect performance.}
    \vspace{-10pt}
    \label{fig:user}
\end{figure}

\textcolor{black}{\noindent \textbf{More details about Eq. (2)}
The two proxies are designed to capture complementary signals from the latent video. Specifically, the appearance proxy aggregates features across time via temporal averaging, and the VFX proxy highlights variations through temporal differences, thereby emphasizing dynamic effect patterns (e.g., particles or flames). By computing frequency-energy indicators from both signals, the model can better distinguish static appearance information from dynamic VFX dynamics for expert routing.}

\textcolor{black}{\noindent \textbf{Actual values of $\sigma_1$ and $\sigma_2$}. The thresholds $\sigma_1$ and $\sigma_2$ are determined according to the latent feature resolution used in FEI computation. In our implementation, their values are $0.46875$ and $0.9375$, respectively.}

\textcolor{black}{\noindent \textbf{More details about FEI and classifier-free guidance.} FEI is only computed on the conditional branch. The conditional and unconditional branches share the same latent input at each timestep under CFG, and thus use the same FEI and routing weights.}

\begin{figure*}[t]
    \centering
    \includegraphics[width=\linewidth]{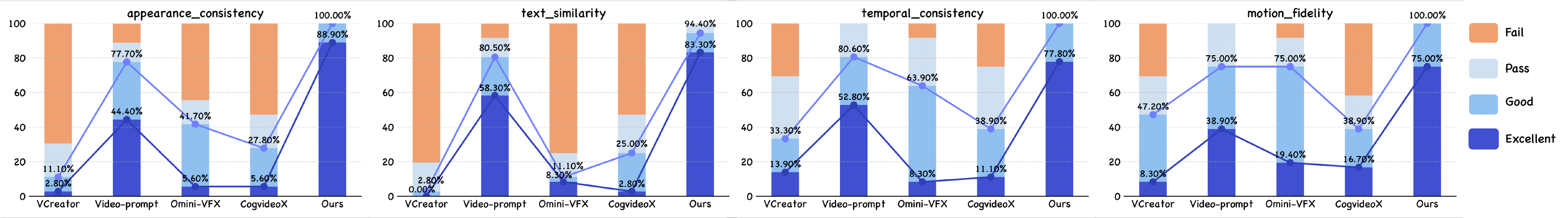}
\caption{\textbf{User study results.} We compare our method against state-of-the-art baselines across four metrics: appearance consistency, text similarity, temporal consistency, and motion fidelity. The results demonstrate that our approach is significantly preferred by human evaluators, consistently achieving the highest ratios of 'Excellent' and 'Good' ratings.}
    \label{fig:user_study}
\end{figure*}

\textcolor{black}{\noindent \textbf{Ablation study about Top-K in our method.} We observe that $k=3$ achieves the best performance across all metrics, while larger $k$ leads to slight degradation, indicating over-segmentation of motion patterns.}

\begin{table}[t]
  \centering
  \caption{\textcolor{black}{\textbf{Ablation on different values of $k$.} 
  We report Text Similarity, Motion Fidelity, Temporal Consistency, and Appearance Consistency. 
  \textcolor{Red}{\textbf{Red}} and \textcolor{Blue}{\textbf{blue}} denote the best and second best results, respectively.}}
  \label{tab:ab_k}
  \resizebox{1.0\linewidth}{!}{%
    \begin{tabular}{l|cccc}
      \toprule
      \multirow{2}{*}{Method}
        & \multicolumn{4}{c}{Quantitative Metrics} \\
      \cmidrule(lr){2-5}
        & Text Sim.$\uparrow$ & Motion Fid.$\uparrow$ & Temp. Cons.$\uparrow$ & App. Cons.$\uparrow$ \\
      \midrule
      $k=2$
        & 0.3015 
        & 0.3745 
        & 0.9413 
        & \textcolor{Blue}{\textbf{0.9373}} \\
      $k=3$ (Ours)
        & \textcolor{Red}{\textbf{0.3274}}
        & \textcolor{Red}{\textbf{0.3985}}
        & \textcolor{Red}{\textbf{0.9522}}
        & \textcolor{Red}{\textbf{0.9425}} \\
      $k=4$
        & \textcolor{Blue}{\textbf{0.3148}} 
        & \textcolor{Blue}{\textbf{0.3862}} 
        & \textcolor{Blue}{\textbf{0.9464}} 
        & 0.9342 \\
      $k=5$
        & 0.3132 
        & 0.3817 
        & 0.9453 
        & 0.9301 \\
      \bottomrule
    \end{tabular}%
  }
\end{table}

\textcolor{black}{\noindent \textbf{Discussion of higher resolutions or longer lengths.} As a flexible framework, it is easy to transfer to stronger video generation backbones. In principle, larger resolutions and longer video sequences can be supported by adopting more capable backbones such as HunyuanVideo-1.5~\cite{kong2025hunyuanvideosystematicframeworklarge}.}

\textcolor{black}{\noindent \textbf{Efficiency comparison with CogVideoX in Fig.3(main paper).} We use the same pretrained CogVideoX weights as the backbone and further finetune it on the OpenVFX dataset for fair comparison.}

\textcolor{black}{\noindent \textbf{Training data.}
The OpenVFX dataset contains 634 annotated VFX videos (480$\times$732) spanning 15 different effect categories, such as \emph{explode}, \emph{melt}, \emph{levitate}, \emph{dissolve}, and \emph{crumble}. Each sample includes text descriptions and instance segmentation masks. We only use 49 frames for Stage~1 training.}

\textcolor{black}{\noindent \textbf{Evaluation of OOD data.} We perform the comparison on the collected 100 out-of-distribution videos. Our method achieves better performance(see the Tab.~\ref{tab:ood}).}

\begin{table}[t]
  \centering
  \caption{\textcolor{black}{\textbf{Evaluation on out-of-distribution (OOD) videos.} 
  We report Text Similarity, Motion Fidelity, Temporal Consistency, and Appearance Consistency on 100 collected OOD videos. 
  $\uparrow$ indicates higher is better. 
  \textcolor{Red}{\textbf{Red}} and \textcolor{Blue}{\textbf{blue}} denote the best and second best results, respectively.}}
  \label{tab:ood}
  \resizebox{1.0\linewidth}{!}{%
    \begin{tabular}{l|cccc}
      \toprule
      Method
        & Text Sim.$\uparrow$ 
        & Motion Fid.$\uparrow$ 
        & Temp. Cons.$\uparrow$ 
        & App. Cons.$\uparrow$ \\
      \midrule
      VFX Creator
        & 0.2291 & \textcolor{Blue}{\textbf{0.3817}} & 0.8669 & 0.8072 \\
      Video-As-Prompt
        & \textcolor{Blue}{\textbf{0.2984}} & 0.3645 & \textcolor{Blue}{\textbf{0.8896}} & \textcolor{Blue}{\textbf{0.8821}} \\
      Omni-Effects
        & 0.2089 & 0.3497 & 0.8213 & 0.7594 \\
      CogVideoX
        & 0.2668 & 0.3689 & 0.8741 & 0.7986 \\
      \midrule
      Ours
        & \textcolor{Red}{\textbf{0.3274}}
        & \textcolor{Red}{\textbf{0.3985}}
        & \textcolor{Red}{\textbf{0.9522}}
        & \textcolor{Red}{\textbf{0.9425}} \\
      \bottomrule
    \end{tabular}%
  }
\end{table}

\textcolor{black}{\noindent \textbf{The details about the loss function.} We would like to clarify that the Freq-constraint loss is not intended to uniquely determine the exact spatial layout of VFX textures or motions by itself. Instead, it serves as a lightweight spectral regularization during test-time adaptation. The structural consistency in our method primarily comes from the frozen I2V backbone, the target-image conditioning, and the Stage-1 frequency-aware MoE priors, which already provide strong spatial and temporal generation bias. Within this framework, the Freq-constraint loss only guides the learnable VFX embedding toward the reference-consistent frequency statistics of the noisy latents, helping the model recover effect-related detail and motion style while preserving the content structure provided by the backbone. Therefore, our claim is not that Eq. (8) alone can specify precise lightning shapes or smoke trajectories, but rather that it provides an effective complementary constraint for improving appearance/detail adaptation under the pretrained generative prior. }

\begin{figure}[t]
    \centering
    \includegraphics[width=\linewidth]{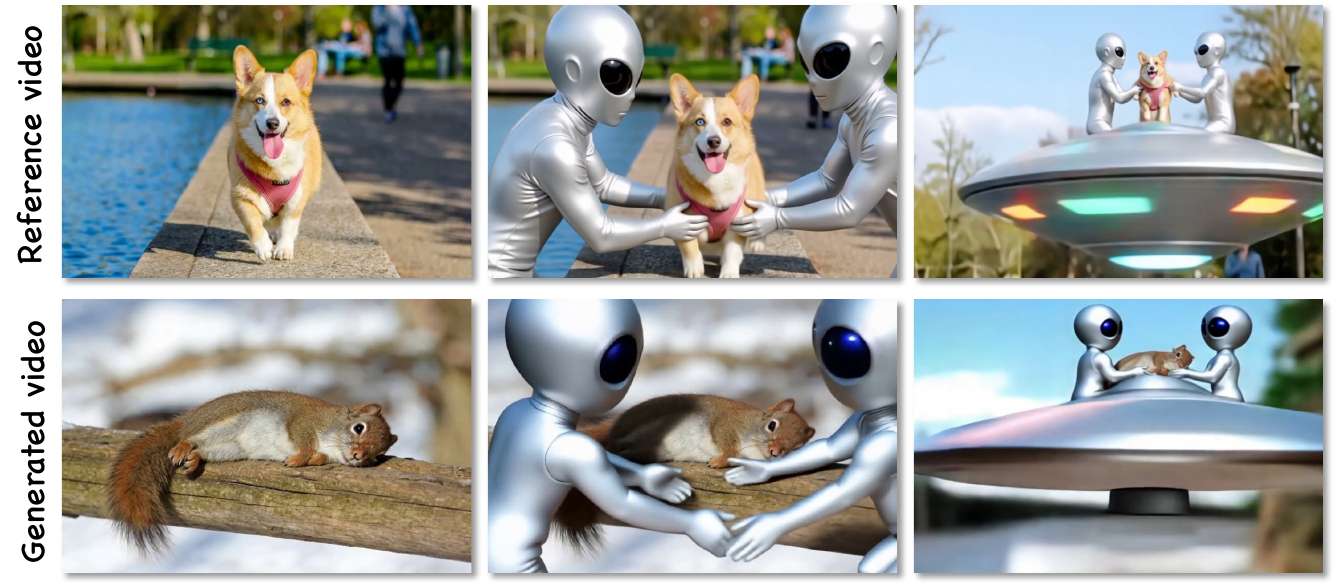}
    \caption{\textbf{Failure cases.} Our method struggles in highly dynamic scenes with large motion or complex interactions, leading to noticeable artifacts and temporal inconsistencies.}
    \vspace{-10pt}
    \label{fig:limitation}
\end{figure}

\textcolor{black}{\noindent \textbf{The details about $\alpha_t$ and $\sigma_t$.}
$\alpha_t$ and $\sigma_t$ are the standard diffusion noise schedule coefficients used in the forward diffusion process. Specifically, $\alpha_t$ controls the scaling of the clean latent $z_0$, while $\sigma_t$ controls the magnitude of the Gaussian noise $\epsilon$ added at timestep $t$. We follow the standard formulation used in latent diffusion models.}

\textcolor{black}{\noindent \textbf{Failure case}. While our method achieves promising results in most cases, it still struggles with very complex visual effects or physically consistent dynamics(shown in Fig.~\ref{fig:limitation}).}

\section{User Study}
To account for the limitations of automatic metrics in capturing human preferences, we conducted a user study with 20 volunteers. 
\textcolor{black}{For each evaluated case, volunteers were shown the target image, the text prompt, the reference effect video, and generated videos from different methods. The method names were hidden, and the generated videos were presented in randomized order to reduce bias.}
\textcolor{black}{We additionally computed inter-annotator agreement using Fleiss' $\kappa$, which is 0.782, indicating substantial agreement among annotators.}

\subsection{User Study Metrics}
\paragraph{Rating dimensions.}
\textbf{User Study:} To account for the limitations of automatic metrics in capturing real-world preferences, we conducted a user study with 20 volunteers. The volunteers categorized each result into one of four levels: excellent, good, pass, or fail, based on motion preservation, appearance diversity, text alignment, and overall quality. The final data is presented in Fig.~\ref{fig:user_study}. Our method outperforms others in both automated metrics and human subjective preferences.
Annotators evaluated each generated result according to the following four criteria:
\begin{itemize}
    \item \textbf{Motion Fidelity:} Does the generated video reproduce motion patterns that are consistent with the reference effect video, without obvious drifting, motion collapse, or unnatural acceleration?
    \item \textbf{Appearance Consistency:} Does the generated visual effect resemble the reference effect in appearance, such as texture, color, and effect components, while remaining coherent across frames?
    \item \textbf{Temporal Consistency:} Is the generated video temporally stable, without obvious flickering, jittering, ghosting, or frame-to-frame structural instability?
    \item \textbf{\textcolor{black}{Text Alignment:}} \textcolor{black}{Judging whether the generated video follows the text prompt, including the main scene content and the desired visual effect described in the prompt. This human rating is different from the automatic CLIP-based Text Similarity metric, and it does not directly measure faithfulness to the specific reference video.}
\end{itemize}

\paragraph{Scoring scale.}
Each dimension was rated using a four-level ordinal scale: 
\textbf{Excellent} (fully satisfies the criterion with no obvious artifacts), 
\textbf{Good} (minor artifacts or deviations), 
\textbf{Pass} (noticeable but acceptable issues), and 
\textbf{Fail} (severe artifacts or failure to satisfy the criterion).

\subsection{User Study Procedure}
\textcolor{black}{Before the formal evaluation, volunteers were given several example cases to familiarize themselves with the rating criteria. During the evaluation, each volunteer independently rated the anonymized and randomly ordered results. The final statistics were obtained by aggregating the ratings across all volunteers and evaluated cases.}

\begin{figure*}[t]
    \centering
    \includegraphics[width=\linewidth]{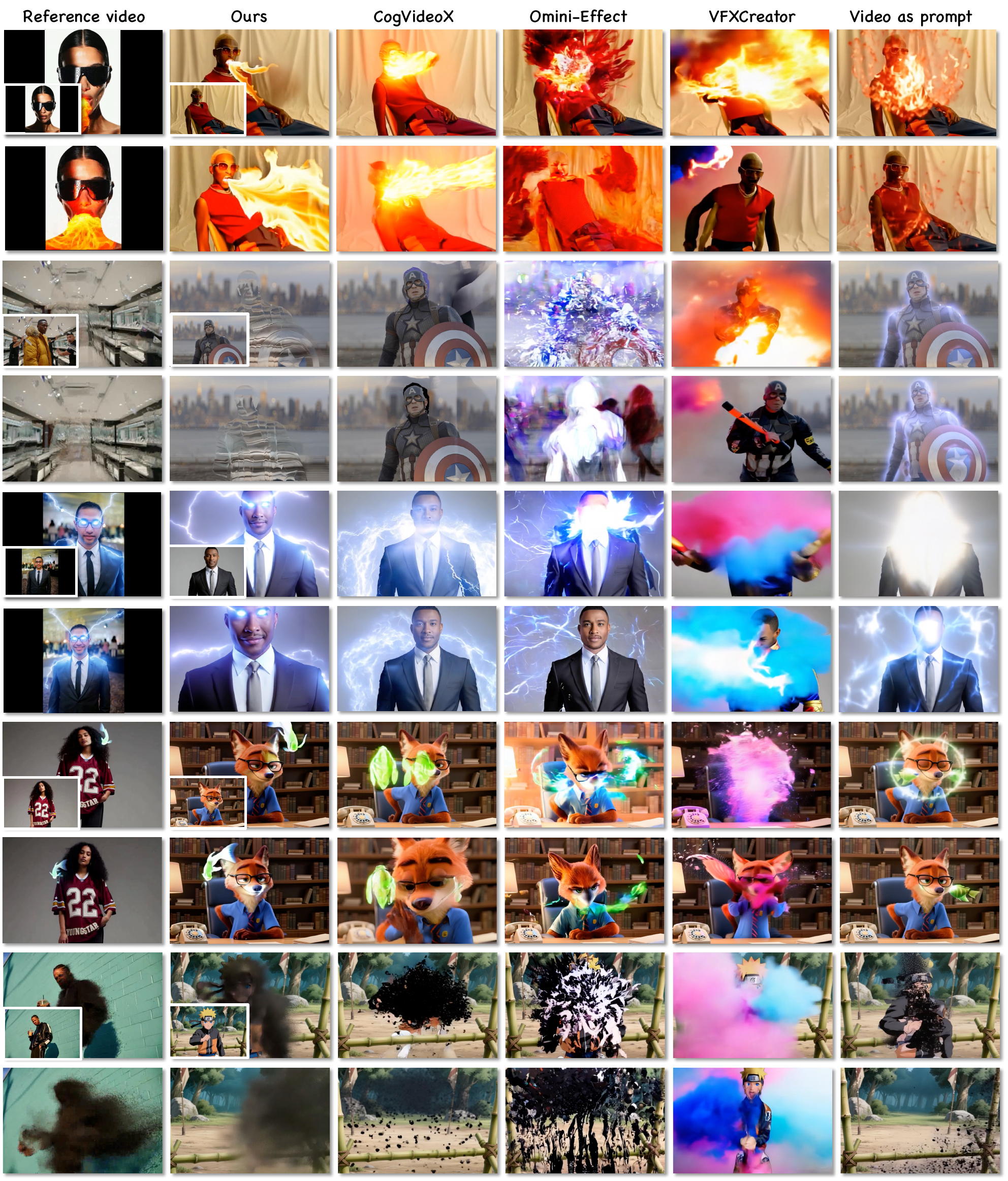}
    \caption{\textbf{More qualitative comparison results with the state-of-the-art methods}.}
    \label{fig:supp_comp}
\end{figure*}

\section{More comparison}
We provide more qualitative results in the Fig.~\ref{fig:supp_comp}. Our method shows a clear advantage in generating high-quality video effects. As shown in the figure, our model can precisely follow the Reference video while keeping the original subject's details. For example, in the first row, our method captures the fire flow more accurately than CogVideoX~\cite{yang2024cogvideox} and VFX Creator~\cite{liu2025vfx}. Moreover, our model preserves the identity of the characters better. In the last row, the facial features and accessories remain sharp, while Omini-Effects~\cite{mao2025omni} creates messy textures. Finally, our method achieves natural lighting and background stability. This proves that our model effectively balances effect transfer and content preservation.

\section{Limitation}

Despite its effectiveness and efficiency, EasyVFX has several limitations.
First, although the proposed Test-Time Training (TTT) strategy is significantly lighter than full model fine-tuning, it still introduces a small amount of optimization overhead at inference time. This additional cost may limit applicability in scenarios with strict real-time or low-latency constraints.

Second, the performance of EasyVFX inherently depends on the representation capacity of the underlying video diffusion backbone. While frequency-aware decoupling improves learning efficiency, the framework may still struggle with effects involving highly complex or non-rigid motion patterns if such dynamics are not well captured by the base model. In addition, EasyVFX does not explicitly resolve potential conflicts between textual guidance and effect-specific visual semantics, which may lead to suboptimal results when textual prompts and reference effects impose competing constraints.

\section{Social potential impact}

EasyVFX aims to reduce the substantial computational and data requirements traditionally associated with high-quality visual effects generation. By enabling effective VFX synthesis under limited training data and modest hardware budgets, our method has the potential to democratize access to professional-grade visual effects technologies. This may benefit independent creators, small studios, educational institutions, and research groups that are often constrained by the cost of large-scale computing resources, thereby fostering broader participation and creativity in digital content production.

From an environmental and sustainability perspective, the proposed frequency-aware design and lightweight training strategy can help reduce energy consumption compared to large-scale, resource-intensive generative models. By emphasizing efficient learning and fast adaptation rather than repeated full-model training, EasyVFX supports more sustainable practices in generative video modeling and deployment.

As with other generative video technologies, there exists a potential risk that visual effects synthesis methods could be misused to create misleading or deceptive media. However, EasyVFX is designed for controlled, reference-based visual effects generation rather than unrestricted video manipulation, which inherently limits its applicability to malicious content creation. Moreover, the method does not introduce new capabilities beyond existing video generation models, but instead focuses on improving efficiency and accessibility.

Overall, EasyVFX contributes toward more inclusive, efficient, and responsible development of generative visual effects technologies, balancing creative empowerment with awareness of potential societal risks.

{
    \small
    \bibliographystyle{ieeenat_fullname}
    \bibliography{sample-bibliography}
}


\end{document}